\documentclass{article}

\usepackage[pdftex]{graphicx, color}
\usepackage{authblk}
\usepackage{appendix}
\usepackage{cite}
\usepackage{url}
\usepackage{ascmac}

\usepackage{amsthm,amsmath,amsfonts,amssymb}
\usepackage{mathabx}
\usepackage{algorithm,algorithmic}
\usepackage{bm}
\usepackage{bbm}
\usepackage{booktabs, multirow}
\usepackage[margin=30truemm]{geometry}
\usepackage[utf8]{inputenc}
\usepackage[T1]{fontenc}
\usepackage{CJKutf8}

\newcommand{\argmin}{\mathop{\rm arg~min}\limits}

\title{Empirical Cumulative Distribution Function Clustering for LLM-based Agent System Analysis}

\author[1]{Chihiro Watanabe}
\author[1]{Jingyu Sun}
\affil[1]{NTT Computer and Data Science Laboratories \\
         3-9-11, Midori-cho, Musashino-shi, Tokyo, Japan\thanks{ch.watanabe@ntt.com}}
\date{}

\begin{document}

\maketitle

\begin{abstract}
Large language models (LLMs) are increasingly used as agents to solve complex tasks such as question answering (QA), scientific debate, and software development. A standard evaluation procedure aggregates multiple responses from LLM agents into a single final answer, often via majority voting, and compares it against reference answers. However, this process can obscure the quality and distributional characteristics of the original responses. In this paper, we propose a novel evaluation framework based on the empirical cumulative distribution function (ECDF) of cosine similarities between generated responses and reference answers. This enables a more nuanced assessment of response quality beyond exact match metrics. To analyze the response distributions across different agent configurations, we further introduce a clustering method for ECDFs using their distances and the $k$-medoids algorithm. Our experiments on a QA dataset demonstrate that ECDFs can distinguish between agent settings with similar final accuracies but different quality distributions. The clustering analysis also reveals interpretable group structures in the responses, offering insights into the impact of temperature, persona, and question topics.
\end{abstract}

\section{Introduction}
\label{sec:introduction}

Following the success of large language models (LLMs), various agent-based approaches ~\cite{Guo24} have been proposed to tackle tasks such as scientific debate ~\cite{Du24, Tang24} and software development ~\cite{Hong24, Qian24, Wu24}. In particular, this paper focuses on the task of question answering (QA), where each question generally has multiple correct answers. Such QA datasets range from multiple-choice questions to test knowledge of models (e.g., CommonsenseQA ~\cite{Talmor19} and SWAG ~\cite{Zellers18} datasets) to questions requiring complex reasoning to answer (e.g., StrategyQA ~\cite{Geva21} and GSM8K ~\cite{Cobbe21} datasets), and are commonly used to measure the task performance of LLM-based agent systems.

A typical evaluation pipeline for LLM-based agent systems involves generating multiple responses per question under a given configuration, then selecting a final answer through decision protocols such as majority voting ~\cite{Kaesberg25, Li24, Wang24moa}.
In such cases, a basic method for comparing multiple settings is to evaluate the consistency between the final answers obtained under each setting and the correct ones. However, such an evaluation criterion alone cannot reveal the tendencies of individual original responses generated by the LLM-based agent. Even if the final answers are the same in a pair of settings, the quality of the original responses may differ. For example, when applying majority voting to $2n+1$ responses, we cannot distinguish between the case where all $2n+1$ responses are correct and the case where only $n+1$ responses are correct. Furthermore, the ``goodness'' of incorrect responses may differ among the original responses. For instance, for the question ``What is the highest mountain in Japan?'', the answer ``Yari-ga-take'' can be considered closer to the correct answer than the answer ``computer,'' even though both are incorrect answers. However, simply measuring the percentage of exact matches with the correct answer does not allow us to distinguish between the quality of these two answers.

\begin{figure*}[t]
\centering
\includegraphics[width=\textwidth]{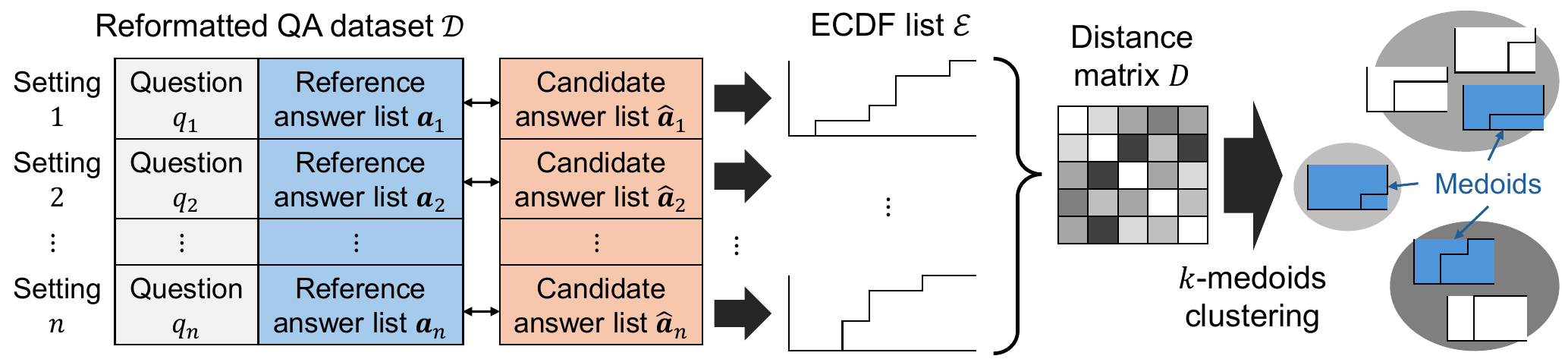}
\caption{The proposed framework of ECDF clustering for LLM-based agent system analysis.}
\label{fig:framework}
\end{figure*}

To solve these problems and obtain more detailed information about the quality of responses given by LLM-based agents to each question, we propose to evaluate a given set of responses based on the empirical cumulative distribution function (ECDF) of their cosine similarities to the correct answers, as shown in Figure \ref{fig:framework}. Using ECDFs to evaluate the responses has two advantages. First, unlike histograms, for which the bin width needs to be determined, there is no need to set any hyperparameters to construct an ECDF. Moreover, by representing the sets of responses as ECDFs, we can make direct comparison across response sets of varying lengths.

One concern with using ECDFs is that there are various types of configurations in LLM-based agent systems, and it is difficult to grasp the overview of the large number of ECDFs corresponding to different settings. Therefore, we propose a method to apply clustering to ECDFs, which allows us to estimate the group structure of ECDFs that are similar to each other. Since ECDFs are different from the typical vector format samples (although a set of cosine similarities for defining an ECDF can be represented as a vector), we propose a new clustering method for ECDFs, each of which corresponds to a set of LLM-based agents' responses in a given setting. Applying clustering to ECDFs has also been proposed in the existing study ~\cite{Kim06} aimed at network traffic anomaly detection, however, their method differs from ours in that the ECDFs are first discretized and converted into vector format samples and then clustered using $k$-means algorithm.

The subsequent part of this paper is organized as follows. In Section \ref{sec:existing_studies}, we review the existing studies on LLM-based agent systems. Next, in Section \ref{sec:proposed}, we propose an ECDF clustering method for analyzing multiple sets of LLM-based agents' responses. In Section \ref{sec:experiments}, we experimentally demonstrate the effectiveness of the proposed ECDF clustering by applying it to two practical QA datasets, changing two types of agent settings (i.e., persona and temperature). Finally, we conclude this paper in Section \ref{sec:conclusion}.


\section{Related Works}
\label{sec:existing_studies}

A typical approach to solving tasks, including QA, with LLM-based agents is to first have each agent generate an answer to a given question, and then combine all the answers to generate a final answer. In this section, we review the existing methods based on such an approach from the following three perspectives: how to generate multiple answers, how to convert multiple answers into a final answer, and how to evaluate the final answer.

First, the generation of multiple answers using LLM-based agents itself involves various settings, including base models of agents, prompt templates, and inter-agent communication styles. Several studies have reported that discussion among LLM-based agents improves performance in reasoning tasks ~\cite{Chen24, Du24, Fang25, Khan24, Liang24}, while others have explored prompting methods to elicit the reasoning ability of models by specifying personas ~\cite{Salewski23, Serapio23} or by describing the reasoning procedure ~\cite{Wei22}. The inter-agent communication structure also affects the cost and quality of the discussion ~\cite{Li24b, Yin23}.

Next, there have been several methods to summarize multiple responses of LLM-based agents into a single final answer. Typical ways to make such decisions are majority or confidence-weighted voting ~\cite{Chan24, Chen24, Kaesberg25, Li24} and introducing an aggregator agent for judging and consolidating the responses ~\cite{Fang25, Liang24, Wang24moa}. However, it has been pointed out that such methods for synthesizing multiple responses eliminate the diversity of the original opinions ~\cite{Wu25}.

Finally, a commonly used criterion for evaluating the final answers for all questions is their average consistency with the correct answers. For example, we can evaluate the quality of the final answer by using exact match, BLEU ~\cite{Papineni02}, and BERTScore ~\cite{Zhang20}. However, when we use these evaluation criteria for the final answers, we might overlook the differences in the distributions of the ``goodness'' of the original answers, as described in Section \ref{sec:introduction}.


\section{ECDF Clustering for LLM-based Agent System Analysis}
\label{sec:proposed}

Let $\tilde{\mathcal{D}} = \{(\tilde{q}_1, \tilde{\bm{a}}_1), \dots, (\tilde{q}_{n_{\mathrm{Q}}}, \tilde{\bm{a}}_{n_{\mathrm{Q}}})\}$ be a QA dataset, where $n_{\mathrm{Q}} \in \mathbb{N}$ is a number of questions. For $i \in [n_{\mathrm{Q}}]$, $\tilde{q}_i \in \mathcal{T}$ is the $i$th question (e.g., $\tilde{q}_i =$``What year is it three years after the year 2000?''), and $\tilde{\bm{a}}_i = [\tilde{a}_{i1}, \dots, \tilde{a}_{in^{(i)}_{\mathrm{R}}}]$ is a list of correct answers (e.g., $\tilde{\bm{a}}_i=[$``2003''$,$ ``The year is 2003.''$]$), where $\mathcal{T}$ is a set of texts, $n^{(i)}_{\mathrm{R}}$ is a number of correct answers for the $i$th question, and $[n] \coloneq \{1, \dots, n\}$.

For each question in QA dataset $\tilde{\mathcal{D}}$, we generate multiple responses using LLM-based agents under mutually different agent settings (e.g., combination of base models, prompts, and temperature). In the subsequent part of this paper, we call a combination of question and agent setting as ``setting'' and it is distinguished from an agent setting, which does not include the setting of a question. Let $\mathcal{D} = \{(q_1, \bm{a}_1), \dots, (q_n, \bm{a}_n)\}$ be the reformatted QA dataset, where $n$ is the number of settings and $q_i$ and $\bm{a}_i$ indicate the question and the correct answer list corresponding to the $i$th setting, respectively. We also define lists $\mathcal{A} = [\bm{a}_1, \dots, \bm{a}_n]$ and $\hat{\mathcal{A}} = [\hat{\bm{a}}_1, \dots, \hat{\bm{a}}_n]$ of reference and candidate answer lists, respectively, where $\hat{\bm{a}}_i = [\hat{a}_{i1}, \dots, \hat{a}_{in^{(i)}_{\mathrm{C}}}]$ is a list of LLM-based agents' responses in the $i$th setting and $n^{(i)}_{\mathrm{C}}$ is its size.

\subsection{Definition of ECDFs}
\label{sec:def_ecdf}

To compare multiple sets of texts (i.e., LLM-based agents' responses) of generally different sizes, we introduce ECDFs of their evaluation values. An ECDF $\hat{F}_{\bm{x}_n}: \mathbb{R} \mapsto [0, 1]$ is a function whose output indicates the proportion of samples in $\bm{x}_n = [x_1, \dots, x_n]$ with an input value or less, and it is given by
\begin{align}
\label{eq:edist}
\hat{F}_{\bm{x}_n}(x) = \frac{1}{n} \sum_{i=1}^n 1_{(-\infty, x]}(x_i).
\end{align}
Here, $1_A: \mathbb{R} \mapsto \{0, 1\}$, $A \subset \mathbb{R}$ is an indicator function, which is given by
\begin{align}
1_A (x) = \begin{cases}
1 & \mathrm{if}\ x \in A,\\
0 & \mathrm{otherwise}.
\end{cases}
\end{align}

Specifically, we propose to define ECDFs based on the cosine similarities between LLM-based agents' responses and correct answers. This allows us to obtain more fine-grained information about the goodness of the answers in each setting, rather than just a binary representation of agreement with a correct answer. By using reformatted QA dataset $\mathcal{D}$ and candidate answers $\hat{\mathcal{A}}$, we define a list $\mathcal{V} = [\bm{v}_1, \dots, \bm{v}_n]$ of cosine similarity lists, where $\bm{v}_i = [v_{i1}, \dots, v_{in^{(i)}_{\mathrm{C}}}]$ is a list of maximum cosine similarities between each LLM-based agent's response and the correct answers in the $i$th setting. For $j \in [n^{(i)}_{\mathrm{C}}]$, we have
\begin{align}
\label{eq:max_cos_sim}
v_{ij} = \max_{a \in \bm{a}_i} g(f_{\bm{\phi}}(a), f_{\bm{\phi}}(\hat{a}_{ij})),
\end{align}
where $f_{\bm{\phi}}: \mathcal{T} \mapsto \mathbb{R}^d$ is a function which embeds input text to $d$ dimensional vector by using an arbitrary model (e.g., neural network) with parameter $\bm{\phi}$ and $g(\bm{u}_1, \bm{u}_2)$ indicates the cosine similarity between vectors $\bm{u}_1$ and $\bm{u}_2$. Based on list $\mathcal{V}$, we define a list $\mathcal{E} = [\hat{F}_{\bm{v}_1}, \dots, \hat{F}_{\bm{v}_n}]$ of ECDFs of cosine similarities.

\subsection{Clustering ECDFs based on Their Distances}
\label{sec:empirical_clustering}

To reveal the group structure and the typical shapes of the ECDFs $\mathcal{E}$ in Section \ref{sec:def_ecdf}, we propose to apply a clustering method to the ECDFs. Specifically, we adopt $k$-medoids clustering, since once a distance matrix is given, it can be applied to the ECDFs in the same way as it is applied to the data vectors.

Let $D=(D_{ij})_{1 \leq i \leq n, 1 \leq j \leq n} \in \mathbb{R}^{n \times n}$ be a distance matrix whose $(i, j)$th entry $D_{ij}$ represents the distance between the ECDFs corresponding to the $i$th and $j$th settings. In this paper, we adopt the following L1 distance\footnote{For $i \in [n]$, let $\mu_i$ be a probability measure on $\mathbb{R}$, which is given by $\mu_i(B)=\frac{1}{n} \sum_{k=1}^{n^{(i)}_{\mathrm{C}}} 1_B (v_{ik})$. This probability measure represents the proportion of samples $\{v_{ik}\}$ that are included in $B \subset \mathbb{R}$. It must be noted that the L1 distance between the $i$th and $j$th ECDFs in Eq. (\ref{eq:d_ij}) is equal to the $L_1$-Wasserstein distance between probability measures $\mu_i$ and $\mu_j$ ~\cite{Barrio99}.}.
\begin{align}
\label{eq:d_ij}
D_{ij} = D_{ji} = \int_{-\infty}^{\infty} \left| \hat{F}_{\bm{v}_i}(x) - \hat{F}_{\bm{v}_j}(x) \right| \mathrm{d}x, \nonumber \\
\mathrm{for\ all}\ 1 \leq i \leq n, 1 \leq j \leq n.
\end{align}
Let $h(\bm{u}_1, \bm{u}_2)$ be the operation that combines vectors $\bm{u}_1$ and $\bm{u}_2$, removes duplicate values, and sorts entries in ascending order (e.g., $h([1, 3, 2], [4, 3, 3, 5]) = [1, 2, 3, 4, 5]$). For all $(i, j)$, we define $\tilde{\bm{v}}_{ij} = h(\bm{v}_i, \bm{v}_j) = [\tilde{v}_{ij1}, \dots, \tilde{v}_{ij\tilde{n}^{(i, j)}_{\mathrm{C}}}]$. Since the difference $| \hat{F}_{\bm{v}_i}(x) - \hat{F}_{\bm{v}_j}(x) |$ between a pair of ECDFs is a step function, which has $(\tilde{n}^{(i, j)}_{\mathrm{C}}-1)$ intervals in the support $[\tilde{v}_{ij1}, \tilde{v}_{ij\tilde{n}^{(i, j)}_{\mathrm{C}}})$, the right side of Eq. (\ref{eq:d_ij}) is given by
\begin{align}
&\int_{-\infty}^{\infty} \left| \hat{F}_{\bm{v}_i}(x) - \hat{F}_{\bm{v}_j}(x) \right| \mathrm{d}x = \nonumber \\
&\sum_{k=1}^{\tilde{n}^{(i, j)}_{\mathrm{C}}-1} \left| \hat{F}_{\bm{v}_i}(\tilde{v}_{ijk}) - \hat{F}_{\bm{v}_j}(\tilde{v}_{ijk}) \right| (\tilde{v}_{ij(k+1)} - \tilde{v}_{ijk}).
\end{align}

Based on distance matrix $D$, we split the ECDFs into $m$ clusters by applying ``Partitioning Around Medoids'' or PAM algorithm ~\cite{Kaufman87} to them. Algorithm \ref{alg:kmedoids} shows the procedure to obtain cluster assignment vector $\bm{c} = [c_1, \dots, c_n] \in [m]^n$ and medoid index vector $\bm{r} = [r_1, \dots, r_m] \in [n]^m$, and it is based on BUILD and SWAP steps. Each entry $c_i$ of cluster assignment vector represents the cluster index of the $i$th ECDF, and each entry $r_i$ of medoid index vector represents the index of the medoid ECDF of the $i$th cluster. In BUILD step, we select initial medoids in a greedy manner so that the sum of distances from each sample to its nearest medoid is as small as possible. Then, in SWAP step, we repeatedly try swapping all pairs of medoid and non-medoid samples and find the best one with the minimum sum of distances from each sample to its nearest medoid. If swapping such an optimal pair leads to a better result than using the original set of medoids, we adopt it. Otherwise, we terminate the algorithm and determine medoid index vector $\bm{r}$ and cluster assignment vector $\bm{c}$ based on the current set of medoids.

\begin{algorithm}[t]
  \caption{PAM algorithm of $k$-medoids clustering}
  \label{alg:kmedoids}
  \begin{algorithmic}
  \renewcommand{\algorithmicrequire}{\textbf{Input: }}
  \renewcommand{\algorithmicensure}{\textbf{Output: }}
    \REQUIRE Distance matrix $D$, number of clusters $m$
    \ENSURE Cluster assignment vector $\bm{c} = [c_1, \dots, c_n] \in [m]^n$, medoid index vector $\bm{r} = [r_1, \dots, r_m] \in [n]^m$
    \STATE \COMMENT{BUILD algorithm for determining initial medoid index vector $\bm{r}$}
    \STATE $r_1 \gets \argmin_{i \in [n]} \sum_{j=1}^n D_{ij}$
    \STATE $J_1 \gets \{r_1\}$
    \FOR{$i=2, \dots, m$}
    \STATE $r_i \gets \argmin_{j \in [n] \setminus J_{i-1}} \sum_{k=1}^n \min_{l \in J_{i-1} \cup \{j\}} D_{kl}$
    \STATE $J_i \gets J_{i-1} \cup \{r_i\}$
    \ENDFOR
    \STATE \COMMENT{SWAP algorithm for improving cluster assignment vector $\bm{c}$ and medoid index vector $\bm{r}$}
    \STATE $J_1 \gets \{r_1, \dots, r_m\}$
    \STATE $J_2 \gets [n] \setminus J_1$
    \STATE $D^* \gets \sum_{k=1}^n \min_{l \in J_1} D_{kl}$
    \WHILE {True}
    \STATE For all $(i, j) \in J_1 \times J_2$, $\tilde{J}_{ij} \gets \left(J_1 \setminus \{i\} \right) \cup \{j\}$
    \STATE For all $(i, j) \in J_1 \times J_2$, $\tilde{D}_{ij} \gets \sum_{k=1}^n \min_{l \in \tilde{J}_{ij}} D_{kl}$
    \STATE $(i^*, j^*) \gets \argmin_{(i, j) \in J_1 \times J_2} \tilde{D}_{ij}$
    \IF{$\tilde{D}_{i^*j^*} < D^*$}
    \STATE $J_1 \gets \tilde{J}_{i^*j^*}$
    \STATE $J_2 \gets [n] \setminus J_1$
    \STATE $D^* \gets \tilde{D}_{i^*j^*}$
    \ELSE
    \STATE \textbf{break}
    \ENDIF
    \ENDWHILE
    \STATE $\bm{r} \gets [r_1, \dots, r_m]$, where $J_1 = \{r_1, \dots, r_m\}$
    \STATE $\bm{c} \gets [c_1, \dots, c_n]$, where $c_i = \argmin_{j \in J_1} D_{ij}$ for all $i \in [n]$
  \end{algorithmic}
\end{algorithm}


\section{Experiments}
\label{sec:experiments}

To verify the effectiveness of the proposed method, we used validation samples of Stanford Question Answering Dataset (SQuAD) ~\cite{Rajpurkar16}. We examined the effects of changing two agent settings, persona and temperature, which we call settings \textbf{P} and \textbf{T}, respectively. For each subject or title, we randomly chose three samples to use in the subsequent procedure. Based on each $i$th original question $q_i$, which is defined as ``context'' and ``question'' values of the dataset concatenated with a space between them, we defined question  $q^*_i = q_i + $`` Just describe your answer in one word without providing any explanation for the answer.'' In all experiments, we set the number of agent settings at $n_0 = 50$ (i.e., the number of settings is $n = n_0 n_{\mathrm{Q}}$). We used GPT-4o mini ~\cite{Hurst24} and paraphrase-MiniLM-L6-v2 ~\cite{Reimers19} as a base model of an LLM-based agent and an embedding model of function $f_{\bm{\phi}}$, respectively, and set the number of responses at $n^{(i)}_{\mathrm{C}} = 10$ for $i \in [n]$.

\begin{itemize}
\item \textbf{Persona settings}: In setting \textbf{P}, we defined the $i$th persona text as $p^*_i = $``You are ''$+ l(p_i) + $`` '', where $l: \mathcal{T} \mapsto \mathcal{T}$ is the operation of converting the first character of the input text to lowercase and the  $p_i \in \mathcal{T}$ is the $i$th sample of ``persona'' subset of Persona Hub dataset ~\cite{Ge24} for $i \in [n_0]$. In setting \textbf{T}, we set $p^*_i =$``'' for all $i \in [n_0]$.
\item \textbf{Temperature settings}: For $i \in [n_0]$, we set the $i$th temperature at $\beta_i = 2(i-1)/n_0$ in setting \textbf{T} and $\beta_i = 1$ in setting \textbf{P}.
\end{itemize}

For each combination of the $i$th question and the $j$th agent settings, we defined the user prompt as $s_{n_0 (i-1) + j} = p^*_j + q^*_i$, input it into the LLM-based agent by setting the temperature at $\beta_j$, and got responses $\hat{\bm{a}}_{n_0 (i-1) + j}$. To allow for variation in the suffix, we removed the suffixes ``.'', ``,'', ``;'', ``:'', and ``\textbackslash\textbackslash'' from both the reference and candidate answers $\mathcal{A}$ and $\hat{\mathcal{A}}$ before evaluation (i.e., final answer selection in Section \ref{sec:preliminary_exp} and computation of maximum cosine similarity in Eq. (\ref{eq:max_cos_sim})).

\subsection{Preliminary Experiment: Diversity of ECDFs}
\label{sec:preliminary_exp}

Before applying the proposed method to the datasets, we examined the following research question: \textit{If the results are similar in terms of accuracy (i.e., the percentage of correct final answers), will their ECDFs end up being similar?} To verify this question, we plotted the ECDFs of the results with the same accuracy together. We determined the final answer in each setting based on the following procedure. First, we define embedding vector list $\mathcal{U}^{(i)} = [\bm{u}_{i1}, \dots, \bm{u}_{in^{(i)}_{\mathrm{C}}})]$ of each $i$th setting, where $\bm{u}_{ij} = f_{\bm{\phi}}(\hat{a}_{ij})$ for $j \in [n^{(i)}_{\mathrm{C}}]$. Next, we select the candidate answer $\hat{a}_{ij^*}$ with the minimum L2 distance from the mean embedding vector $\bar{\bm{u}}_i$.
\begin{align}
j^* = \argmin_{j \in [n^{(i)}_{\mathrm{C}}]} \|\bm{u}_{ij} - \bar{\bm{u}}_i\|_2, \ \ \ \bar{\bm{u}}_i = \frac{1}{n^{(i)}_{\mathrm{C}}} \sum_{j=1}^{n^{(i)}_{\mathrm{C}}} \bm{u}_{ij}.
\end{align}
By using the above procedure, we can always determine a final answer even if all the candidate answers are different from each other. Finally, we define a binary value $b_i \in \{0, 1\}$ representing the correctness of each $i$th setting. We define $b_i=1$ if there exists a reference answer $a \in \bm{a}_i$ that matches the final answer $\hat{a}_{ij^*}$ and $b_i=0$ otherwise. The accuracy of each $k$th subject is given as the mean value of $b_i$ for all $i$ that correspond to the $k$th subject.

\begin{figure*}[t]
\centering
\includegraphics[width=0.8\textwidth]{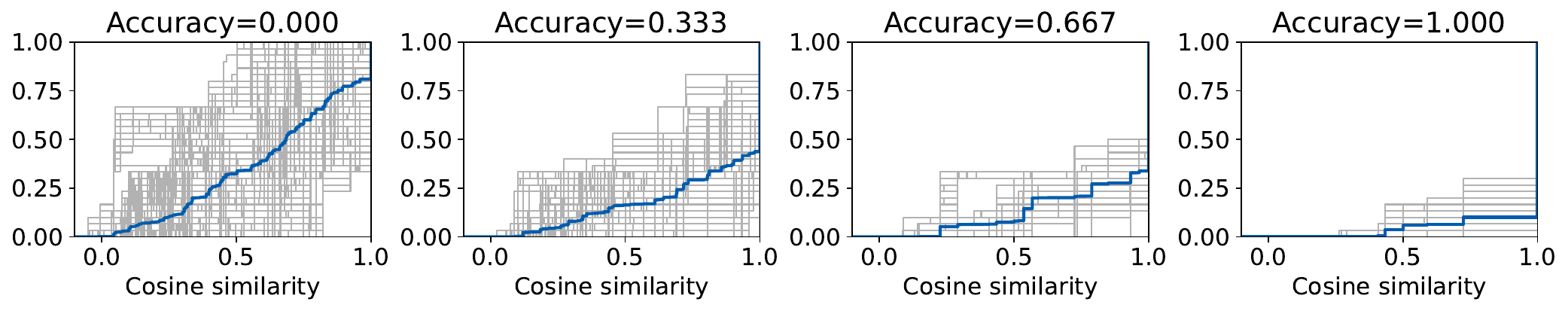}
\caption{ECDFs for each value of accuracy in setting \textbf{P}. Blue line shows the centroid.}
\label{fig:edists_sp}
\includegraphics[width=0.8\textwidth]{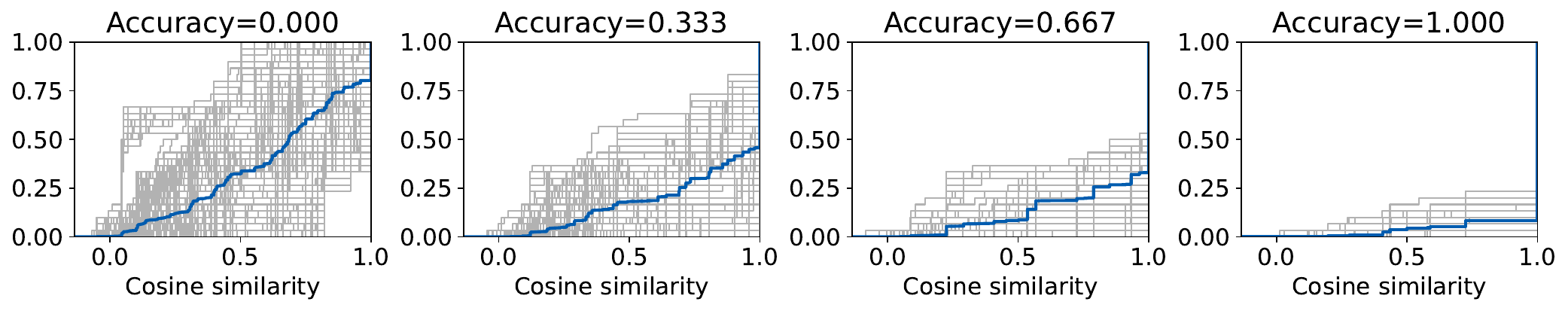}
\caption{ECDFs for each value of accuracy in setting \textbf{T}.}
\label{fig:edists_st}
\end{figure*}

Figs. \ref{fig:edists_sp} and \ref{fig:edists_st} show the ECDFs for each value of accuracy in settings \textbf{P} and \textbf{T}, respectively. It must be noted that there were only four possible accuracy values (i.e., $0$, $1/3$, $2/3$, $1$), since the accuracy was measured as the percentage of correct answers for three questions of the same subject. For $i \in [4]$, the $i$th centroid represents an ECDF of the concatenated samples corresponding to the $i$th accuracy value. From these results, we see that even with the same accuracy, the ECDFs can vary greatly, especially when the accuracy is low. Moreover, when the accuracy is one, not all the responses match the correct answers. That is, there were cases where the original responses contained wrong answers, however, the final answer matched the correct answer through the final answer selection process.

\subsection{Application to SQuAD Dataset}

We applied ECDF clustering in Section \ref{sec:empirical_clustering} to the candidate answers for SQuAD dataset by setting the number of clusters at $m=16$. Before plotting the clustering results, we redefined the cluster indices so that the clusters with smaller indices indicate better results. Specifically, we defined matrix $W = (W_{ij})_{1 \leq i \leq m, 1 \leq j \leq m} \in \{0, 1\}^{m \times m}$, whose $(i, j)$th entry is given by
\begin{align}
W_{ij} = 1_{(-\infty, 0)} \left\{ \int_{-\infty}^{\infty} \left[ \hat{F}_{\bm{v}_{r_i}}(x) - \hat{F}_{\bm{v}_{r_j}}(x) \right] \mathrm{d}x \right\}, \nonumber \\
\mathrm{for\ all}\ 1 \leq i \leq m, 1 \leq j \leq m.
\end{align}
Based on matrix $W$, we defined the number of ``wins'' of each $k$th cluster as $w_k = \sum_{j=1}^m W_{kj}$ and redefined the cluster indices so that the $k$th cluster corresponds to the $k$th largest value of $\{w_k\}$.

For visibility, we plotted the cluster assignments as matrix $C = (C_{ij})_{1 \leq i \leq n_{\mathrm{Q}}, 1 \leq j \leq n_0} \in [m]^{n_{\mathrm{Q}} \times n_0}$, whose $(i, j)$th entry represents the cluster index corresponding to the combination of the $i$th question and the $j$th agent setting. Instead of plotting matrix $C$ as is, we applied matrix reordering to it so that the rows and columns with mutually similar values are closer together. Specifically, we applied multidimensional scaling (MDS) ~\cite{Rodgers92, Spence74} to the rows and columns of matrix $C$, obtained their one-dimensional features, and sorted them according to the feature values.

\begin{figure*}[p]
\centering
\includegraphics[width=0.75\textwidth]{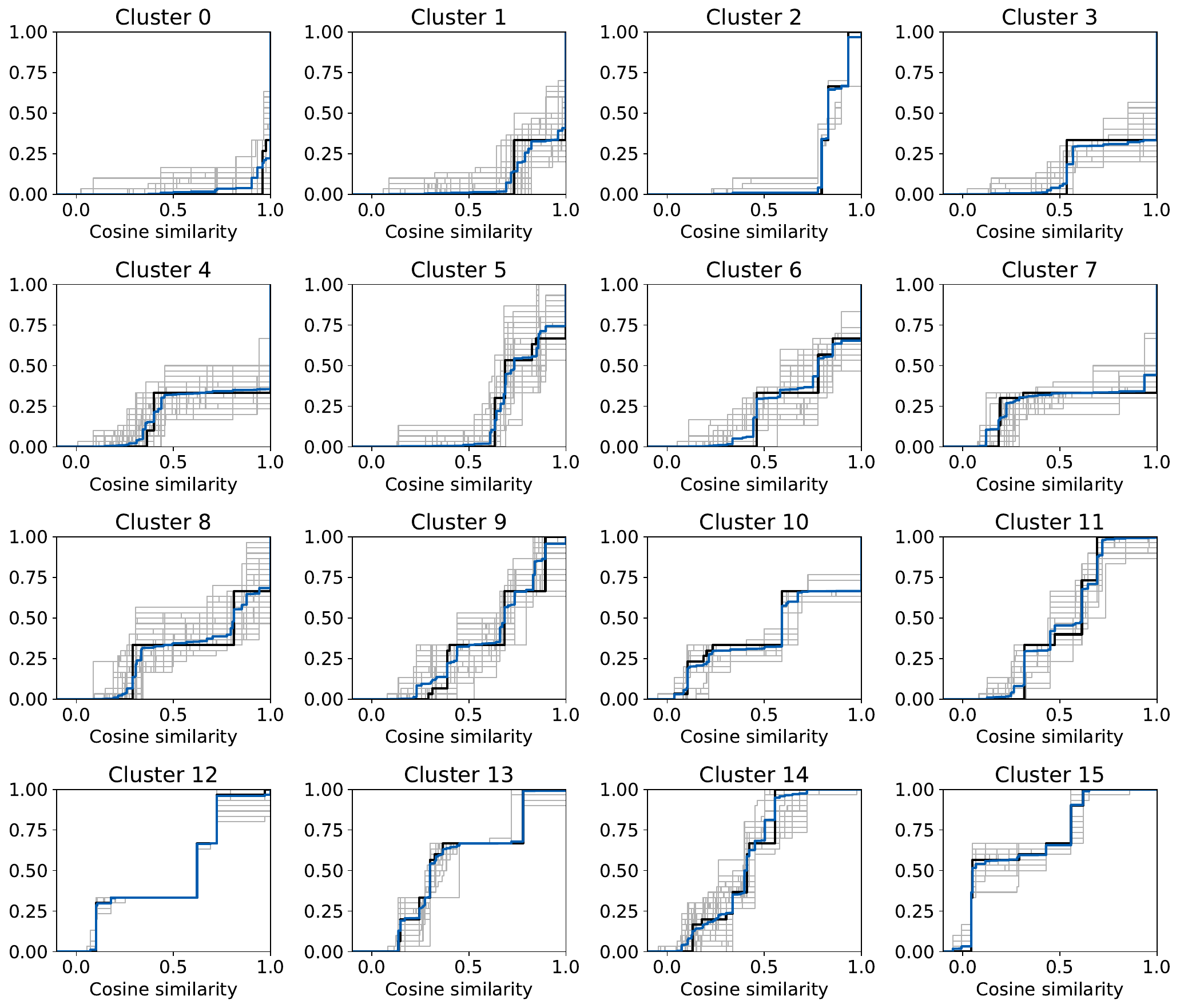}
\caption{ECDFs of each cluster in setting \textbf{P}. Black and blue lines show the medoid and centroid of the cluster, respectively.}
\label{fig:cluster_sp}
\includegraphics[width=0.85\textwidth]{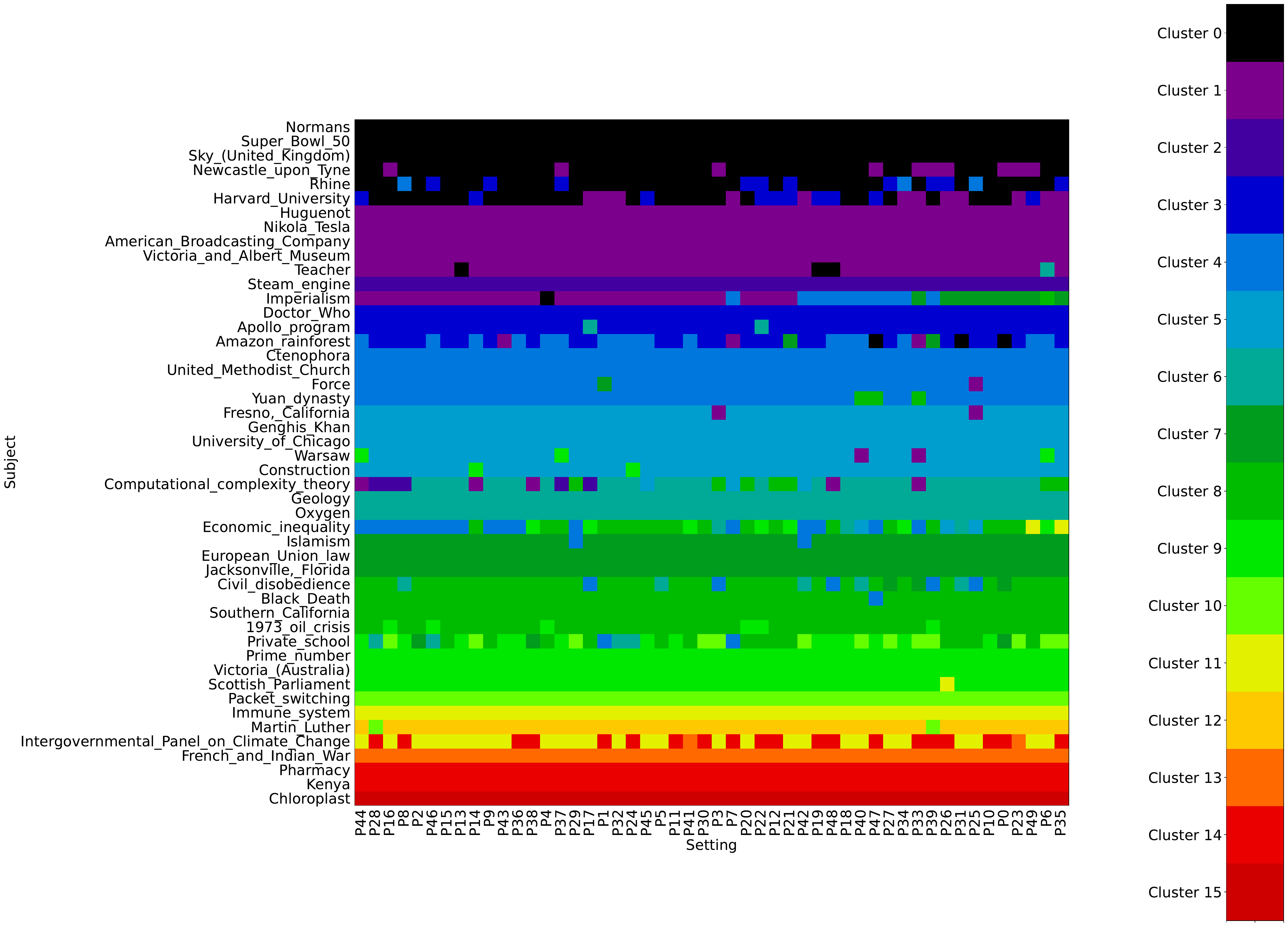}
\caption{Cluster assignments in setting \textbf{P}.}
\label{fig:assign_sp}
\end{figure*}

\begin{figure*}[p]
\centering
\includegraphics[width=0.75\textwidth]{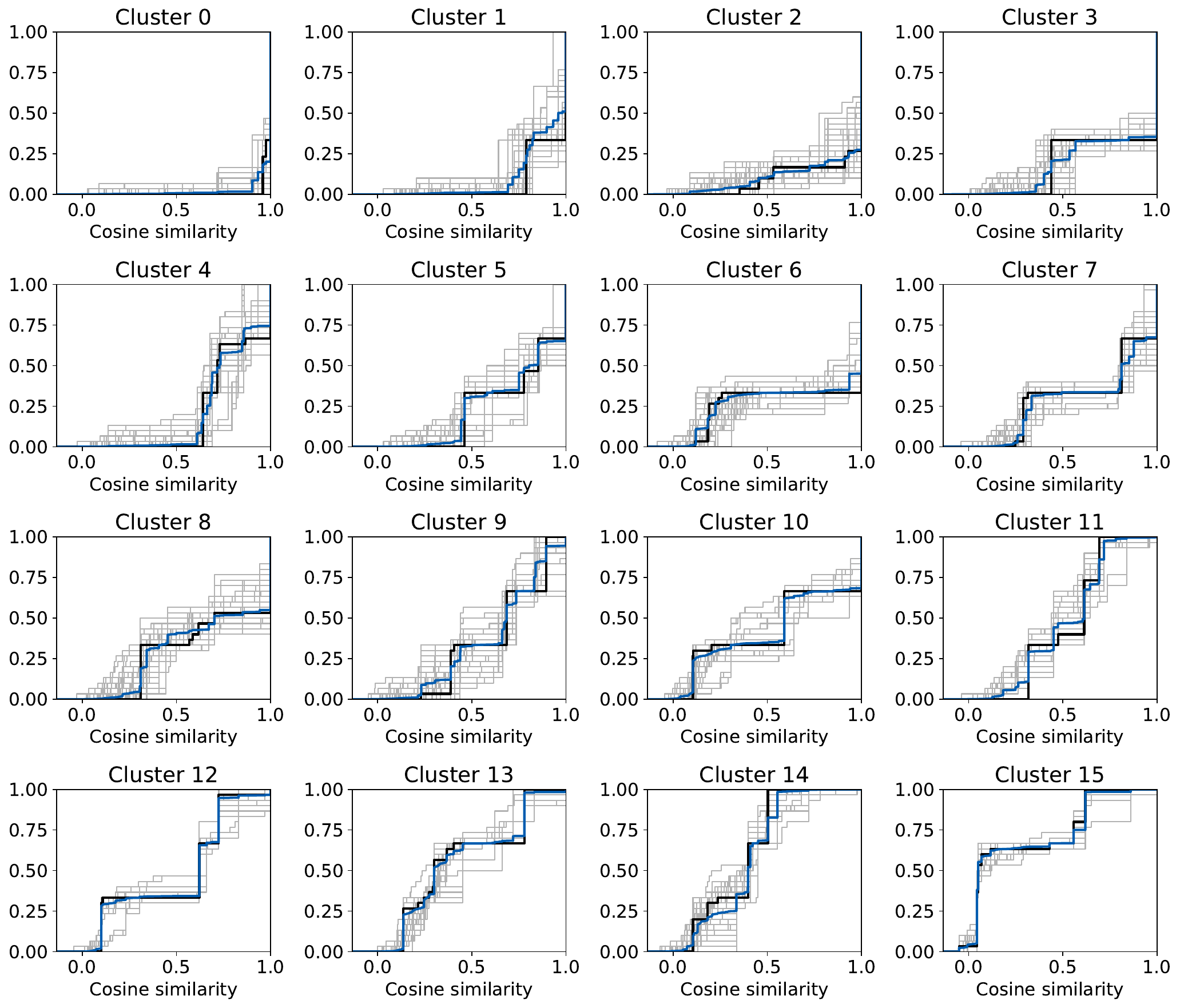}
\caption{ECDFs of each cluster in setting \textbf{T}.}
\label{fig:cluster_st}
\includegraphics[width=0.85\textwidth]{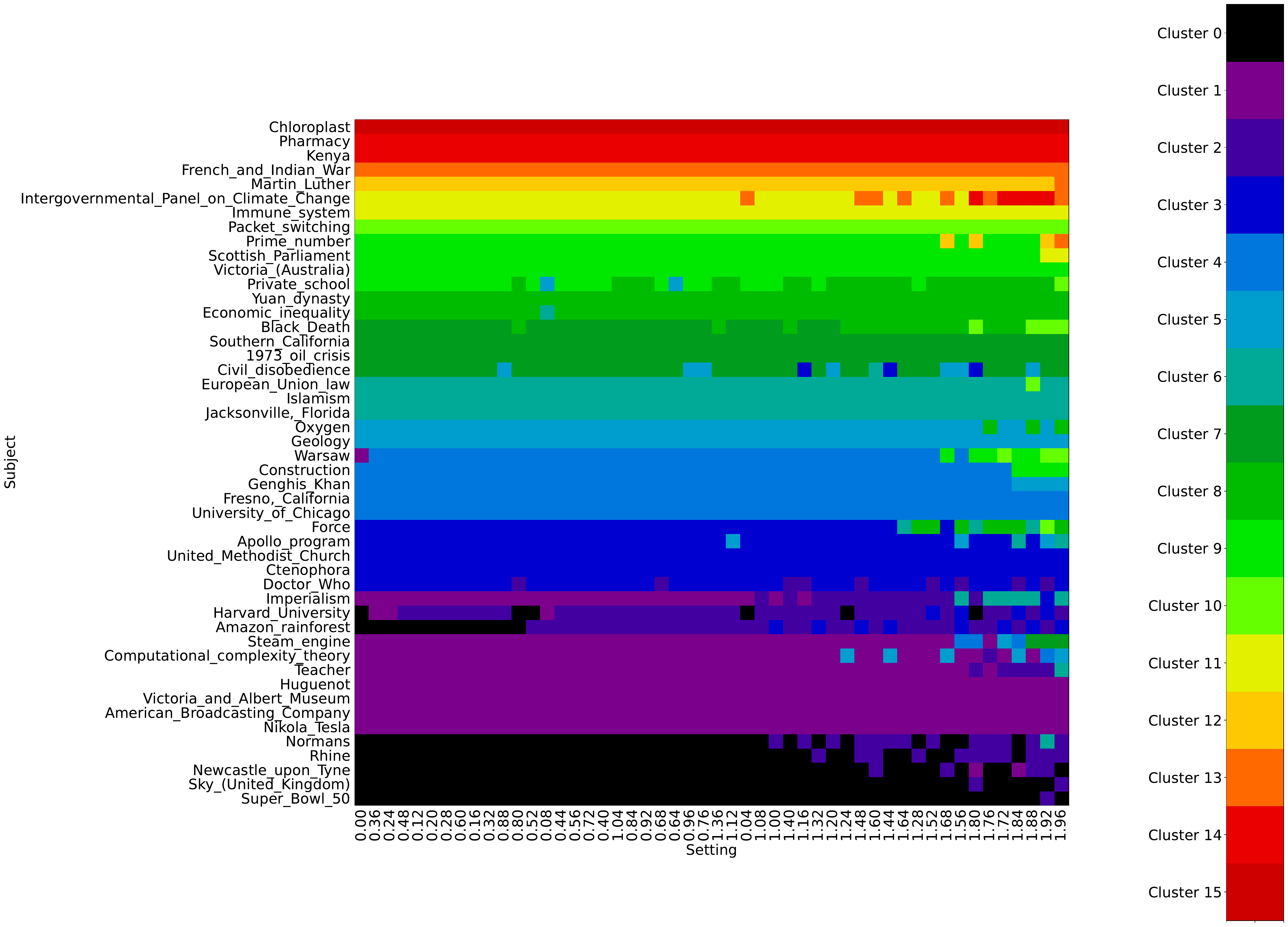}
\caption{Cluster assignments in setting \textbf{T}.}
\label{fig:assign_st}
\end{figure*}

Figs. \ref{fig:cluster_sp} and \ref{fig:cluster_st} show the ECDFs of each cluster in settings \textbf{P} and \textbf{T}, respectively, and Figs. \ref{fig:assign_sp} and \ref{fig:assign_st} show the cluster assignments in settings \textbf{P} and \textbf{T}, respectively. As in Section \ref{sec:preliminary_exp}, each $i$th centroid represents an ECDF of the concatenated samples in the $i$th cluster. From Figs. \ref{fig:cluster_sp} and \ref{fig:cluster_st}, we see that relatively similar ECDFs were assigned to each cluster compared to the results in Section \ref{sec:preliminary_exp}. Fig. \ref{fig:assign_sp} shows that there was little difference in the ECDFs depending on the persona setting, however, the ECDFs changed significantly depending on the subject of the question. For example, questions in subject ``Normans'' and ``Super-Bowl-50'' were relatively easy for agents to answer correctly, while questions in subject ``Chloroplast'' often resulted in answers that were far from the correct answers. Fig. \ref{fig:assign_st} shows that the temperature also did not significantly affect the ECDFs, and that the cluster structure at low temperatures collapsed when the temperature was set relatively high.

\begin{figure*}[t]
\centering
\includegraphics[width=0.8\textwidth]{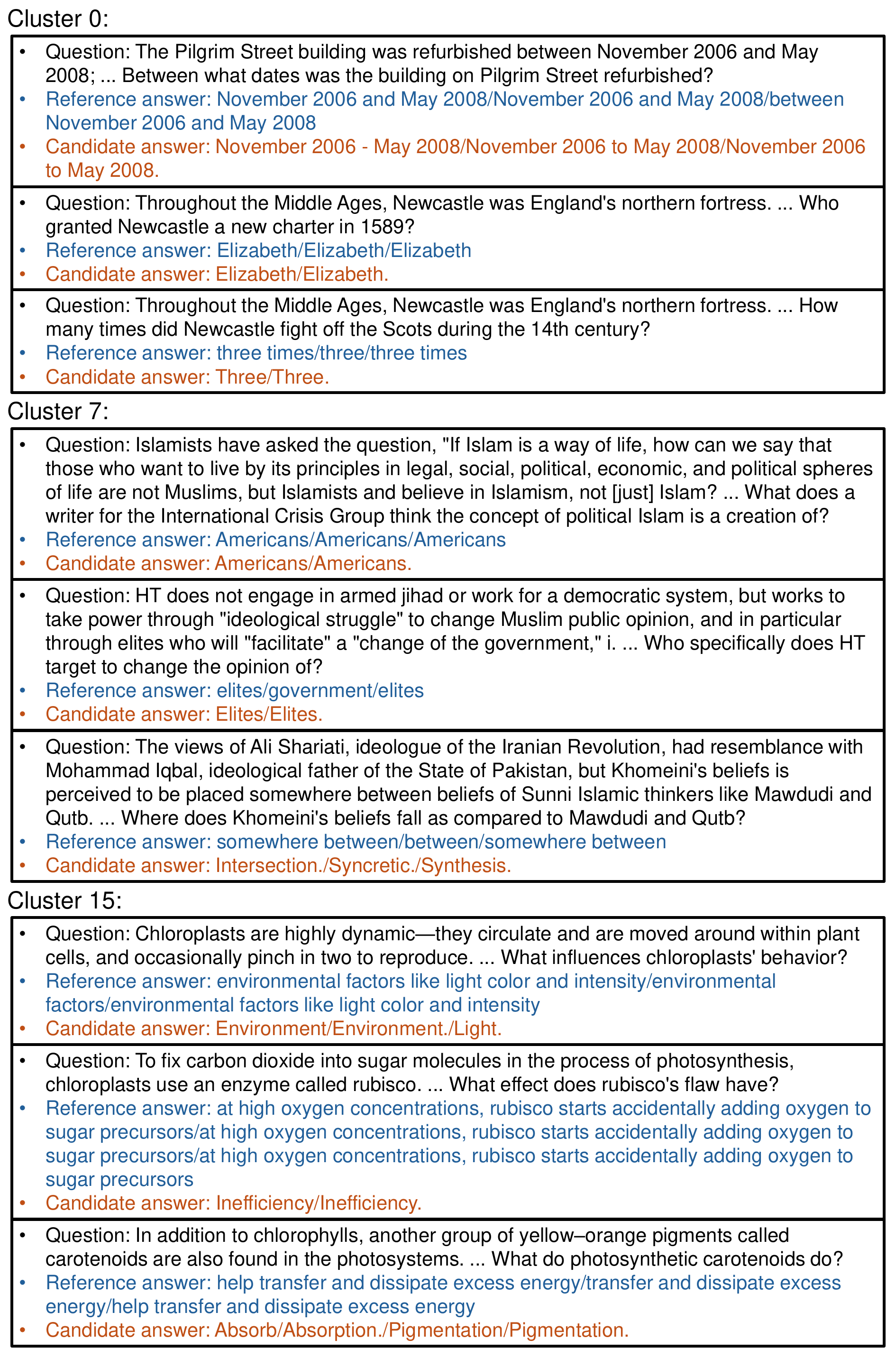}
\caption{Example answers of medoids of Clusters $0$, $7$, and $15$ in setting \textbf{P}. Regarding the candidate answers, only unique answers are listed.}
\label{fig:example_sp}
\end{figure*}

As additional information for each cluster, we show example answers of medoids of Clusters $0$, $7$, and $15$ in Figure \ref{fig:example_sp}. From this figure, we see that medoids of Clusters $0$ and $7$ correspond to the questions whose correct answers are expressed in relatively simple words, while that of Cluster $15$ corresponds to the questions that require more complex and longer sentences, leading to the worst ECDFs.


\section{Conclusion}
\label{sec:conclusion}

In this paper, we developed a new method for evaluating the responses generated by LLM-based agents. For each setting, we first define a single ECDF that represents the cumulative distribution of similarities between the generated answers and the correct ones, and then estimate the cluster structure of the ECDFs corresponding to all the settings. We showed the effectiveness of the proposed method by applying it to a practical dataset. The proposed method successfully revealed the cluster structure of the ECDFs, each of which corresponds to a combination of a temperature/persona and a question topic.


\bibliographystyle{unsrt}
\bibliography{ecdf}

\end{document}